\documentclass[10pt,journal,compsoc]{IEEEtran}
\usepackage{graphicx}
\usepackage{makecell}
\usepackage{tabularx}
\usepackage{ulem}
\usepackage{color,xcolor}
\usepackage{amsmath}
\usepackage{ulem}   % 期刊

\ifCLASSOPTIONcompsoc
  \usepackage[nocompress]{cite}
\else
  \usepackage{cite}
\fi

\begin{document}
\title{A Survey of Knowledge-Enhanced Pre-trained Language Models}
%
%
% author names and IEEE memberships
% note positions of commas and nonbreaking spaces ( ~ ) LaTeX will not break
% a structure at a ~ so this keeps an author's name from being broken across
% two lines.
% use \thanks{} to gain access to the first footnote area
% a separate \thanks must be used for each paragraph as LaTeX2e's \thanks
% was not built to handle multiple paragraphs
%
%
%\IEEEcompsocitemizethanks is a special \thanks that produces the bulleted
% lists the Computer Society journals use for "first footnote" author
% affiliations. Use \IEEEcompsocthanksitem which works much like \item
% for each affiliation group. When not in compsoc mode,
% \IEEEcompsocitemizethanks becomes like \thanks and
% \IEEEcompsocthanksitem becomes a line break with idention. This
% facilitates dual compilation, although admittedly the differences in the
% desired content of \author between the different types of papers makes a
% one-size-fits-all approach a daunting prospect. For instance, compsoc 
% journal papers have the author affiliations above the "Manuscript
% received ..."  text while in non-compsoc journals this is reversed. Sigh.

\author{Linmei Hu,~\IEEEmembership{}
        Zeyi Liu,~\IEEEmembership{}
        Ziwang Zhao,~\IEEEmembership{}
        Lei Hou, ~\IEEEmembership{}
        Liqiang Nie, \textit{Senior Member, IEEE}
        and Juanzi Li ~\IEEEmembership{}

\IEEEcompsocitemizethanks{
\IEEEcompsocthanksitem L. Hu is with the School of Computer Science and Technology,  Beijing Institute of Technology, Beijing 100081, China. (e-mail: hulinmei@bit.edu.cn).\\
\IEEEcompsocthanksitem Z. Liu and Z. Zhao are with the School of Computer Science, Beijing University of Posts and Telecommunications, Beijing 100876, China. (e-mail: liu\_zy@bupt.edu.cn, zhaoziwang@bupt.edu.cn).\\
\IEEEcompsocthanksitem L. Hou is with the Department of Computer Science and Technology, Tsinghua University, Beijing 100084, China. (e-mail: houlei@tsinghua.edu.cn).\\
\IEEEcompsocthanksitem L. Nie is with the School of Computer Science and Technology, Harbin Institute of Technology (Shenzhen) (e-mail: nieliqiang@gmail.com).\\
\IEEEcompsocthanksitem J. Li is with the Department of Computer Science and Technology, Tsinghua University, Beijing 100084, China. (e-mail:lijuanzi@tsinghua.edu.cn).\\
}% <-this % stops a space
\thanks{}
}

% The paper headers
% \markboth{IEEE TRANSACTIONS ON KNOWLEDGE AND DATA ENGINEERING}%
% {Shell \MakeLowercase{\textit{et al.}}: Bare Advanced Demo of IEEEtran.cls for IEEE Computer Society Journals}
% The only time the second header will appear is for the odd numbered pages
% after the title page when using the twoside option.
% 

% for Computer Society papers, we must declare the abstract and index terms
% PRIOR to the title within the \IEEEtitleabstractindextext IEEEtran
% command as these need to go into the title area created by \maketitle.
% As a general rule, do not put math, special symbols or citations
% in the abstract or keywords.
\IEEEtitleabstractindextext{%
\begin{abstract}

    Pre-trained Language Models (PLMs) which are trained on large text corpus via self-supervised learning method, have yielded promising performance on various tasks in Natural Language Processing (NLP). However, though PLMs with huge parameters can effectively possess rich knowledge learned from massive training text and benefit downstream tasks at the fine-tuning stage, they still have some limitations such as poor reasoning ability due to the lack of external knowledge. Research has been dedicated to incorporating knowledge into PLMs to tackle these issues. In this paper, we present a comprehensive review of Knowledge-Enhanced Pre-trained Language Models (KE-PLMs) to provide a clear insight into this thriving field. We introduce appropriate taxonomies respectively for Natural Language Understanding (NLU) and Natural Language Generation (NLG) to highlight these two main tasks of NLP. For NLU, we divide the types of knowledge into four categories: linguistic knowledge, text knowledge, knowledge graph (KG), and rule knowledge. The KE-PLMs for NLG are categorized into KG-based and retrieval-based methods. Finally, we point out some promising future directions of KE-PLMs.

\end{abstract}

% Note that keywords are not normally used for peerreview papers.
}

% make the title area
\maketitle

% To allow for easy dual compilation without having to reenter the
% abstract/keywords data, the \IEEEtitleabstractindextext text will
% not be used in maketitle, but will appear (i.e., to be "transported")
% here as \IEEEdisplaynontitleabstractindextext when compsoc mode
% is not selected <OR> if conference mode is selected - because compsoc
% conference papers position the abstract like regular (non-compsoc)
% papers do!
\IEEEdisplaynontitleabstractindextext
% \IEEEdisplaynontitleabstractindextext has no effect when using
% compsoc under a non-conference mode.

% For peerreview papers, this IEEEtran command inserts a page break and
% creates the second title. It will be ignored for other modes.
\IEEEpeerreviewmaketitle

\ifCLASSOPTIONcompsoc
\IEEEraisesectionheading{\section{Introduction}\label{sec:introduction}}
\else
\section{Introduction}
\label{sec:introduction}
\fi

\IEEEPARstart
{W}{ith} the continuous development of deep learning technologies in recent years, Pre-trained Language Models (PLMs) which are trained with unsupervised objectives on massive text corpora, have been widely used in the field of Natural Language Processing (NLP), and yielded state-of-the-art performance on various downstream tasks. Different from traditional supervised learning, PLMs based on self-supervised learning are usually pre-trained on general-purpose large-scale unlabeled data first and then fine-tuned on small-scale labeled data for the specific tasks. Representative work, such as BERT \cite{devlin2018bert}, GPT \cite{radford2018improving}, T5 \cite{raffel2020exploring}, has refreshed benchmark records constantly in many Natural Language Understanding (NLU) and Natural Language Generation (NLG) tasks, successfully promoting the development of NLP.

As the size of PLMs grows larger, PLMs with hundreds of millions of parameters have been extensively demonstrated to possess the ability of capturing rich linguistic \cite{liu2019linguistic,belinkov2019analysis,vulic2020probing} and factual knowledge \cite{petroni2019language,li2022pre} in certain probings. However, due to the lack of explicit representation of knowledge in the raw data, PLMs suffer from limited performance on downstream tasks.  In particular,  
the prior study has found that traditional pre-training objectives often have weak {symbolic reasoning capabilities} \cite{alon2020olmpics} since PLMs tend to concentrate on the word co-occurrence information. 
Incorporating knowledge into PLMs can empower their memorization and reasoning \cite{sun2021ernie}.
For instance, in the language understanding problem of "The monument to the people's Heroes sits solemnly on [MASK] square", traditional PLMs predict the output of the masked position as "the", while knowledge-enhanced PLMs predict "Tiananmen", which is more accurate. 

% Although they can capture rich language information from text sentence corpus and generate accurate language texts, almost all of them ignore knowledge information and thereby fail to generate output towards capturing the human commonsense
% 原先是 KG-BART \cite{liu2021kg} 中的原句（上面注释掉的这句）
For language generation, although existing PLMs are able to obtain rich language information from text corpus and generate correct sentences, almost all of them fail to generate output towards capturing the human commonsense, since they overlook the external world knowledge \cite{liu2021kg}. 
In other words, sentences generated by PLMs often conform to the grammatical norm, but not to logic. For example, given a concept set \{hand, sink, wash, soap\} to generate a sentence, conventional PLMs may generate { "hands washing soap on the sink"}, while the PLM with extra knowledge generates "man is washing his hands with soap in a sink", which is more natural and logical.

\begin{figure*}[htbp]
        \centering
        \includegraphics[width=0.98\linewidth]{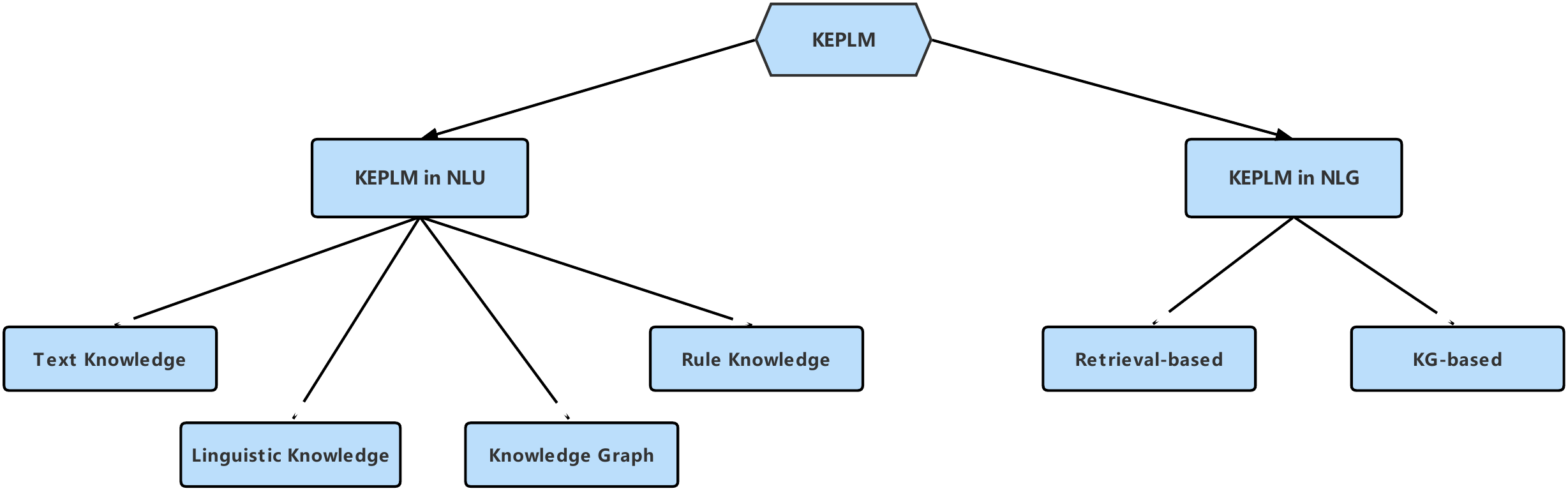}
        \caption{Taxonomy of Knowledge Enhanced Pre-trained Language Models (KE-PLMs) based on the two core tasks of NLP: Natural Language Understanding (NLU) and Natural Language Generation (NLG).}
        \label{1}
    \end{figure*}

% \citet{Knowledge Enhanced Pretrained Language Models: A Compreshensive Survey}
To address the above issues, explicitly incorporating knowledge into PLMs has been an emerging trend in recent NLP studies. Wei et al. \cite{wei2021knowledge} reviewed the knowledge-enhanced PLMs along three taxonomies: types of knowledge sources, knowledge granularity, and application. 
% 添加了一个PLMKE综述
Yin et al. \cite{yin2022survey} summarized the recent progress of pre-trained language model-based knowledge-enhanced models (PLMKEs) according to three crucial elements of them: knowledge sources, knowledge-intensive NLP tasks, and knowledge fusion methods. 
In this work, considering the fact that injecting knowledge into language models can promote both NLU and NLG tasks, and these two areas have different focuses,  
we aim to present a comprehensive review of Knowledge-Enhanced Pre-trained Language Models (KE-PLMs) in the two areas to provide respective insights of KE-PLMs in NLU and NLG. The main contributions of this survey can be summarized as follows:
	
\begin{itemize}
\item[(1)] In this survey, we divide KE-PLMs into two main categories according to the downstream tasks: NLU and NLG. Appropriate taxonomies are respectively presented to highlight the focuses of these two different kinds of tasks in NLP.

\item [(2)] For NLU, KE-PLMs are further divided into four sub-categories according to the types of knowledge: linguistic knowledge, text knowledge, knowledge graph (KG), and rule knowledge. For NLG, focused on the knowledge sources, KE-PLMs are further categorized into retrieval-based methods and KG-based methods. Fig. \ref{1} shows our proposed taxonomies for NLU and NLG.
    
 \item[(3)] We discuss some possible directions that may tackle the existing problems and challenges of KE-PLMs in the future.
\end{itemize}

The rest of this paper is arranged as follows. In section 2, we provide the background of PLMs under the development of training paradigms in NLP. In Section 3, we introduce the taxonomy of KE-PLMs in the field of NLU. In Section 4, we introduce the taxonomy of KE-PLMs in the field of NLG. For both NLU and NLG fields, we discuss the representative work of each leaf category in the taxonomy. In Section 5, we propose the possible research directions of KE-PLMs in the future based on the existing limitations and challenges. Finally, we conclude in Section 6.
	
    % scale=0.7

%-----------------------section 2 KE-PLM in NLU---------------------------------

\section{Background}
    \textbf{\textit{Pre-trained Language Models.}} Although the idea of pre-training on a language modeling task is not novel, the training paradigm has shifted to \textit{pre-train and fine-tune}, with the emergence of ELMo \cite{Peters2018ELMo} and ULMFiT \cite{Howard2018ULMFiT}. Both are based on Long Short-Term Memory (LSTM) architecture. They propose to fine-tune the language model layer by layer for downstream tasks, and their performance demonstrates the competitiveness of pre-trained language models.
    
    \textcolor{black}{Unlike the early fully supervised method that learns salient features from limited data \cite{liu2021self}, language models can be trained on a large amount of raw textual data to obtain general-purpose representations. Then, the pre-trained models will be applied to various downstream tasks by fine-tuning them through task-specific objective functions \cite{liu2021pre}. For example, UNILM \cite{Dong2019UNILM} unifies three language modeling objectives, which can be adapted to NLU and NLG tasks simultaneously.}

    As the Transformer architecture with multi-head self-attention mechanism is put forward \cite{VaswaniSPUJGKP17},  all popular language models, including GPT \cite{radford2018improving}, BERT \cite{devlin2018bert}, BART \cite{Lewis2020BART} and T5 \cite{raffel2020exploring}  are proposed based on the Transformer. The multi-head self-attention mechanism allows every word to attend to each other, making the models capture long-range dependencies and learn more expressive representations. These models differ in the model structure and training objectives. In particular, GPT is an autoregressive language model (unidirectional) which predicts the next word given all the previous words. BERT is a masked language model (bidirectional) that aims to predict the ``masked" word conditioned on all the other words. BART and T5 are encoder-decoder language models that learn to generate a sequence when given an input sequence.

    \textbf{\textit{Prompt Learning.}} Instead of adapting PLMs to different downstream tasks by designing specific objective functions, “\textit{pre-train, prompt, and predict}” which reformulates downstream tasks through textual prompts has taken the place of “pre-train, fine-tune” to become the fourth paradigm in NLP \cite{liu2021pre}. In this paradigm, there is even no need for fine-tuning models using task-specific training objectives, PLMs themselves can directly be employed to predict the output that is desired, breaking through the problem of data constraints and bridging the gap of objective forms between pre-training and fine-tuning \cite{petroni2019language, shin2020autoprompt, schick2020exploiting, li2021prefix, gu2021ppt, zhang2021differentiable, ding2021openprompt}. Previous work \cite{liu2021pre} conducts a comprehensive survey on the emerging field of prompt-based learning. 
    
    Though this prompt learning method has achieved promising results via constructing prompt information without changing the structure and parameters of PLMs significantly, it also calls for the necessity of choosing the most appropriate prompt template and verbalizer \cite{schick2020exploiting} which may have a great impact on model's performance \cite{schick2020exploiting}. To this end, some work \cite{han2021ptr, li2021sentiprompt, hu2021knowledgeable, chen2022knowprompt, ye2022ontology} has proposed to use knowledge as prompt to enhance the prompt-tuning process and reduce the cost of template construction and label mapping. Through injecting domain/task-relevant knowledge at the fine-tuning stage, PLMs can better serve downstream tasks and obtain better performance. In this survey, we also investigated the knowledge-enhanced pre-trained language models that incorporate external knowledge via prompt learning.

%-----------------------section 2 KE-PLM in NLU---------------------------------

\section{KE-PLMs for NLU}
    
    NLU is a subpart of NLP concerning all the methods which enable machines to understand and interpret the content of textual data. It extracts core semantic information from the unstructured text and applies this information to downstream tasks, thus playing a vital role in applications such as text classification, relation extraction, named entity recognition (NER), and dialogue system. In line with the taxonomy shown in Fig. \ref{1}, we divide the knowledge incorporated by KE-PLMs which are designed for NLU tasks into the following four categories according to their different types, that is, linguistic knowledge, text knowledge, knowledge graph, and rule knowledge. For each category, we discuss its representative methods.

\subsection{Incorporating Linguistic Knowledge into PLMs}
	
	Linguistic knowledge, mainly divided into lexical knowledge and syntax tree, is the most common auxiliary feature incorporated into PLMs \cite{yu2022survey}. Among them, lexical knowledge includes but is not limited to Part-of-Speech (POS) tagging, and sentiment tags of words. LIBERT \cite{lauscher2019specializing} introduces lexical relation classification (LRC) as a new pre-training task on the basis of standard BERT objectives, using synonyms and hypernym-hyponym pairs to predict whether two words are in specific semantic relations, which enhances the modeling ability of PLMs for semantic information. SenseBERT \cite{levine2019sensebert} integrates word-supersense (e.g., noun.food, noun.state) and predicts their corresponding supersenses by restoring the masked words, which can explicitly learn the semantic information of words in a given context. SKEP \cite{tian2020skep} improves the effect of PLM on the sentiment analysis task by integrating sentiment knowledge (sentiment words, polarity, and aspect-sentiment pairs). Sentiprompt \cite{li2021sentiprompt} incorporates sentiment knowledge about aspects, opinions, and polarities into prompt through the construction of consistency and polarity judgment templates, and explicitly models term relations between aspect and opinion terms, better introducing task-related knowledge for the language models through prompt-tuning methods. LET \cite{lyu2021let} integrates semantic information of HowNet to improve the Chinese sentence matching task. KEAR \cite{xu2021human} combines the knowledge from ConceptNet, dictionary entry definition, and labeled training data to enhance its performance on commonsense knowledge question answering. DictBERT \cite{chen2022dictbert} takes dictionary knowledge as the external source, and achieves knowledge enhancement for pre-training tasks by means of contrastive learning.
	
\begin{figure*}[htbp]
        \centering
        \includegraphics[width=0.8\textwidth]{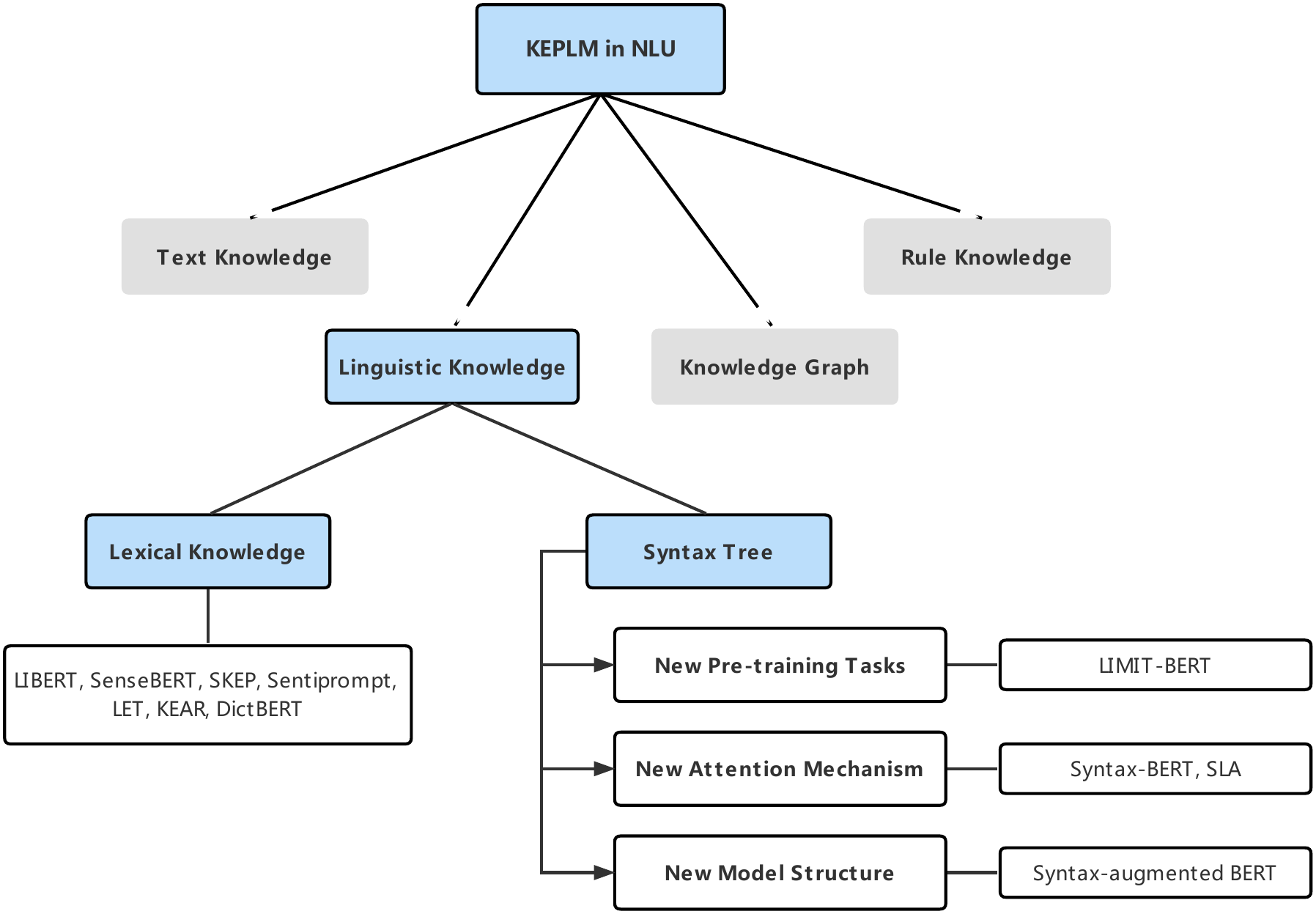}
        \caption{Categorization of Linguistic knowledge.}
        \label{2}
    \end{figure*}

    Considering the different ways of incorporating syntax tree knowledge into PLMs, we divided the KE-PLMs into three categories as illustrated in Fig. \ref{2}, including introducing new relevant pre-training tasks \cite{zhou2019limit}, adopting new attention mechanism \cite{bai2021syntax, li2020improving}, and designing new model structure \cite{sachan2020syntax}.
    For instance, LIMIT-BERT \cite{zhou2019limit} realizes multi-task learning across five linguistic tasks, and simply sums up these task-specific losses together on the basis of a variety of linguistic knowledge such as POS tagging, semantic role labeling (SRL), dependency relations, syntax trees, and etc. in model training. Syntax-BERT \cite{bai2021syntax} provides additional syntax information by constructing a syntax tree parser, and generates a new attention mechanism according to the constructed syntax tree. \textcolor{black}{Syntax-augmented BERT \cite{sachan2020syntax} introduces a syntax-based graph neural network to fuse the syntax information from dependency trees to improve PLM.}

    Despite the above methods consistently incorporating linguistic knowledge into PLMs, they differ in the stages of knowledge fusion. LIBERT \cite{lauscher2019specializing}, SenseBERT \cite{levine2019sensebert}, SKEP \cite{tian2020skep}, Dictbert \cite{chen2022dictbert}, LIMIT-BERT \cite{zhou2019limit}, and Syntax-BERT \cite{bai2021syntax} fuse linguistic knowledge in the pre-training stage of PLMs to enhance the representation of the input, while Sentiprompt \cite{li2021sentiprompt}, LET \cite{lyu2021let}, and KEAR \cite{xu2021human} fuse knowledge in the fine-tuning stage of PLMs for improving task performance.
	
\subsection{Incorporating Text Knowledge into PLMs}
	
	Text knowledge is usually retrieved from {general-domain} text collection (such as WikiText \cite{khandelwal2019generalization}, Wiktionary \cite{xu2021human}) or large corpus (such as Wikipedia \cite{guu2020retrieval}). KNN-LM \cite{khandelwal2019generalization} selects the nearest K neighbors from training samples as knowledge incorporated into PLM, and its earlier idea comes from cache-LM \cite{grave2016improving} which remains the first K words in the cache. REALM \cite{guu2020retrieval} utilizes text corpus to train a text retriever explicitly, exploiting information retrieved from external knowledge bases such as Wikipedia documents to help the prediction of tokens that are masked. ExpBERT \cite{murty2020expbert} and KEAR \cite{xu2021human} also incorporate textual descriptions into their models to improve performance.
    \textcolor{black}{OK-Transformer \cite{cui2022enhancing} incorporates large-scale out-of-domain commonsense descriptions to enhance the representation of input text.}
    Kformer \cite{yao2022kformer} obtains some external text knowledge through retrieval, and injects this knowledge into the FFN layer of Transformer. REINA \cite{wang2022training} retrieves some training samples similar to the input from external datasets as knowledge to enhance PLM. UniK-QA \cite{oguz2020unik} and UDT-QA \cite{ma2022open} use text, knowledge graph and table knowledge together, and transform all the knowledge into text for knowledge enhancement. 
	
	% 原来在 rule knowledge 下面，改到这里了
    In addition to the type of general text knowledge that has been mentioned above, some work also utilizes domain-specific corpora or scientific and technical literature for pre-training tasks. In particular, BioBERT \cite{lee2020biobert} and SciBERT \cite{beltagy2019scibert} 
    \textcolor{black}{conduct the pre-training process on large-scale scientific domain corpora and achieve promising results on downstream academic NLP tasks.}
    \textcolor{black}{S2ORC-BERT \cite{lo2019s2orc} leverages the same method as SciBERT on a larger corpus which contains numerous academic papers covering dozens of academic disciplines, and slightly promotes its performance on several downstream tasks.}
    
    As for the stage of fusing  knowledge, REALM \cite{guu2020retrieval}, BioBERT \cite{lee2020biobert}, SciBERT \cite{beltagy2019scibert} and S2ORC-BERT \cite{lo2019s2orc} integrate text knowledge in the pre-training stage, while KNN-LM \cite{khandelwal2019generalization}, ExpBERT \cite{murty2020expbert}, KEAR \cite{xu2021human}, OK-Transformer \cite{cui2022enhancing}, Kformer \cite{yao2022kformer}, REINA \cite{wang2022training}, UniK-QA \cite{oguz2020unik}, and UDT-QA \cite{ma2022open} integrate knowledge in the fine-tuning stage.

\subsection{Incorporating Knowledge Graph into PLMs}
	
	Knowledge graph can be considered as a kind of powerful expression that represents real world knowledge in the structural form of graph, where its nodes represent entities, and edges represent relations between entities \cite{hogan2021knowledge,xue2022knowledge}. Compared with other types of knowledge, such as text knowledge which is unstructured, knowledge graph often contains more abundant structured information which  makes it more applicable to enhance the potential learning capability of models \cite{wang2017knowledge,wang2021context,fang2022knowledge,li2022oerl}. Here, we further divide KG into entity knowledge and triplet knowledge as shown in Fig. 3.
	
\subsubsection{Entity Knowledge}
	
	As the most basic semantic unit, entity plays a vital role in many natural language scenarios, such as machine reading comprehension, NER, sentiment analysis, and etc \cite{zhao2022connecting}. Incorporating entity knowledge into PLMs helps to improve the semantic understanding ability of models and their performance in downstream tasks as well. According to the different ways of incorporating entity knowledge, we divide them into three sub-categories as follows.
 %we divide them from three perspectives according to the main contribution of the method.}
	
	One is to design entity related pre-training tasks \textcolor{black}{\cite{sun2019ernie}, \cite{xiong2019pretrained}, \cite{wang2022kecp}, \cite{qin2020erica}, \cite{zhang2022dkplm}}. ERNIE \cite{sun2019ernie} proposes a multi-stage knowledge masking strategy of both entity level and phase level. WKLM \cite{xiong2019pretrained} integrates knowledge of external entities through the entity replacement strategy, and determines whether each entity is replaced by performing binary classification prediction. KECP \cite{wang2022kecp} adopts contrastive learning and prompt learning to integrate entity knowledge. 
	
	The second is to change the attention mechanism of the model. For example, \textcolor{black}{LUKE \cite{yamada2020luke} introduces an entity-aware self-attention mechanism to capture the entity information in calculating attention scores. }

	The third is to change the model structure \cite{zhang2019ernie, peters2019knowledge, fevry2020entities, yu2022jaket}. ERNIE-THU \cite{zhang2019ernie} leverages an N-layer T-encoder structure which is similar to the BERT-base model to extract text information, and then uses a proposed K-encoder to integrate entity knowledge.
    KnowBert \cite{peters2019knowledge} retrieves entity embeddings related to the input text via an entity linker and then updates contextual word representations through word-to-entity attention,  allowing the long-range interactions between words and all entities.
    \textcolor{black}{EaE \cite{fevry2020entities} expands each entity mention embedding in the input text by fusing it with the top K relevant entity embeddings retrieved by a proposed component called an entity memory layer.}

    % realize interactive calculation by using attention between entity spans, which inserts entity information into each token more explicitly.

    Note that several methods may belong to more than one sub-category, and we assign them to the sub-category according to their most significant contributions.
	
    \begin{figure*}[htbp]
        \centering
        \includegraphics[width=\textwidth]{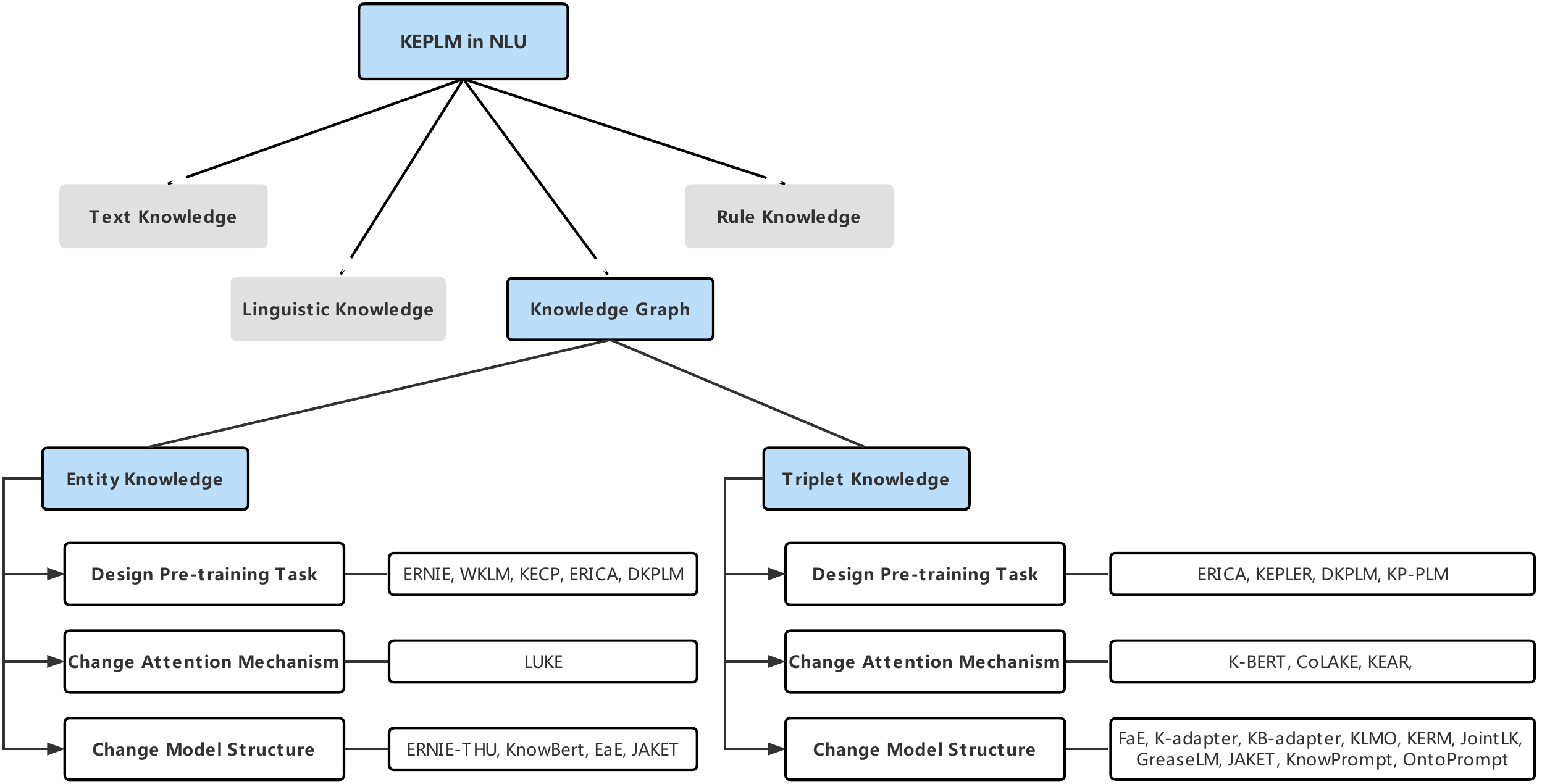}
        \caption{Categorization of knowledge graph and injection methods of each sub-category.}
        \label{3}
    \end{figure*}
	
\subsubsection{Triplet Knowledge}
	
    In addition to the entity knowledge mentioned above, there are also a great number of relational triples in the knowledge graph, which can provide sufficient structured information for PLMs and also improve the semantic understanding ability of them. Similar to the KE-PLMs incorporating entity knowledge, we also divide KE-PLMs in this category into three sub-categories according to their specific ways of knowledge incorporation.  %\textcolor{red}{Notably, several methods may belong to more than one category, and we assign the category according to their most significant contributions.}
    
    One is to design pre-training tasks related to triplets \textcolor{black}{\cite{qin2020erica}, \cite{wang2021kepler}, \cite{zhang2022dkplm}, \cite{wang2022knowledge})}. 
    ERICA \cite{qin2020erica} introduces both entity and relation discrimination tasks to deepen PLM's understanding of entities and relations through contrastive learning.
    KEPLER \cite{wang2021kepler} trains knowledge embedding and masked language modeling objectives jointly to improve the knowledge representation. DKPLM \cite{zhang2022dkplm} focuses on long-tail entities, enriching semantic information of low-frequency entities with knowledge graphs. KP-PLM \cite{wang2022knowledge} designs two knowledge-aware pre-training tasks to incorporate knowledge triplets into multiple continuous prompts for NLU tasks.
    In Table 1, we summarize entity and triplet relevant pre-training tasks designed by the existing work.
    
    \begin{table}[]
		\caption{Entity/Triplet related pre-training tasks. Here MLM represents masked language modeling.}
		\label{3}
		\setcellgapes{2pt}
		\makegapedcells
		\resizebox{\linewidth}{!}{%
			\begin{tabular}{l|l}
				\hline
				\makecell[l]{Method} & Pre-training Tasks/Objectives \\
				\hline
				ERNIE \cite{sun2019ernie} & Token-level, Phrase-level and Entity-level MLM \\
				\hline
                WKLM \cite{xiong2019pretrained} & Entity replacement, MLM \\
                \hline
                KECP \cite{wang2022kecp} & \makecell[l]{Token-level MLM, Span-level contrastive learning} \\
                \hline
                ERICA \cite{qin2020erica} & Entity and relation discrimination tasks, MLM \\
                \hline
                DKPLM \cite{zhang2022dkplm} & \makecell[l]{Relational knowledge decoding, Token-level MLM} \\
                \hline
                KP-PLM \cite{wang2022knowledge} & \makecell[l]{Prompt relevance inspection, Masked prompt modeling} \\
                \hline
                KEPLER \cite{wang2021kepler} & Knowledge embedding, MLM \\
				\hline
			\end{tabular}
		}
	\end{table}
    
    The second is to change the attention mechanism of the model \cite{liu2020k}, \cite{sun2020colake}, \cite{xu2021human}). \textcolor{black}{K-BERT \cite{liu2020k} leverages a knowledge layer to inject relevant triplets from KG into the input sentence and transform it into a knowledge-rich sentence tree. Then this sentence tree is converted into a visible matrix to control the visible area of each word in the sentence, preventing the sentence from deviating from the correct semantics due to injecting too much knowledge.}
    If K-BERT intends to expand input text into a sentence tree, the core concept of CoLAKE \cite{sun2020colake} is to expand the input context into word-knowledge graphs (WK graphs), and then feed these constructed WK graphs into masked self-attention to gather information of nodes. KEAR \cite{xu2021human} proposes an external attention mechanism to enhance the Transformer architecture, and integrate external knowledge into its prediction process. We illustrate their attention mechanism in Table 2, where $ \boldsymbol{Q} $, $ \boldsymbol{K} $, $ \boldsymbol{V} $ denote the query, key, and value matrices, respectively. $ d_{k} $ is the dimension of the key, which is used as the scaling factor.

    % In order to overcome the knowledge noise (KN) issue that a sentence may deviate from its correct semantic when there is too much knowledge incorporated, 
    % \textcolor{red}{K-BERT \cite{liu2020k} introduces soft-position embedding and visible matrix into its model. Through injecting triplets from KG into the input sentence and transforming it into a knowledge-rich sentence tree, the embedding of a word can only come from context of the same branch, while the words of different branches cannot affect each other.} 
    
    \begin{table*}[]
		\caption{Examples of changing attention mechanism for incorporating triplet knowledge. }
		\label{3}
		\setcellgapes{2pt}
		\makegapedcells
		\resizebox{\textwidth}{!}{%
			\begin{tabular}{|l|l|l|l|}
				\hline
				\makecell[l]{Method} & Attention Mechanism & Formalized expression & Variation in Calculation \\
				\hline
				% LUKE \cite{yamada2020luke} & Entity-aware Self-Attention & $ $ & \\
				% \hline
				K-BERT \cite{liu2020k} & \textcolor{black}{Masked Self-Attention} & $ Attn(\boldsymbol{Q, K, V}) = Softmax(\frac{\boldsymbol{QK^\top} + \boldsymbol{M}}{\sqrt{d_{k}}})\boldsymbol{V} $ & \makecell[l]{visible matrix $ \boldsymbol{M} $ for the input sentence tree, \\$ M_{ij} = 0 $ if token $ i $ and token $ j $ are in the \\same branch, while $ M_{ij} = -\infty $ if not} \\
				\hline
				CoLAKE \cite{sun2020colake} & Masked Self-Attention & $ Attn(\boldsymbol{Q, K, V}) = Softmax(\frac{\boldsymbol{QK^\top}}{\sqrt{d_{k}}} + \boldsymbol{M})\boldsymbol{V} $ & \makecell[l]{mask matrix $ \boldsymbol{M} $ for word-knowledge graph, \\$ M_{ij} = 0 $ if node $ i $ and node $ j $ are connected, \\while $ M_{ij} = -\infty $ if not} \\
				\hline
				KEAR \cite{xu2021human} & External Attention & $ Attn(\boldsymbol{Q, K, V}) = Softmax(\frac{\boldsymbol{QK^\top}}{\sqrt{d_{k}}})\boldsymbol{V} $ & \makecell[l]{concatenate extra knowledge $ \boldsymbol{K} $ into the input \\text $ \boldsymbol{H}_{0} = [\boldsymbol{X};\boldsymbol{K}] $} \\
				\hline
			\end{tabular}
		}
	\end{table*}

    \begin{figure*}[htbp]
        \centering
        \includegraphics[width=\textwidth]{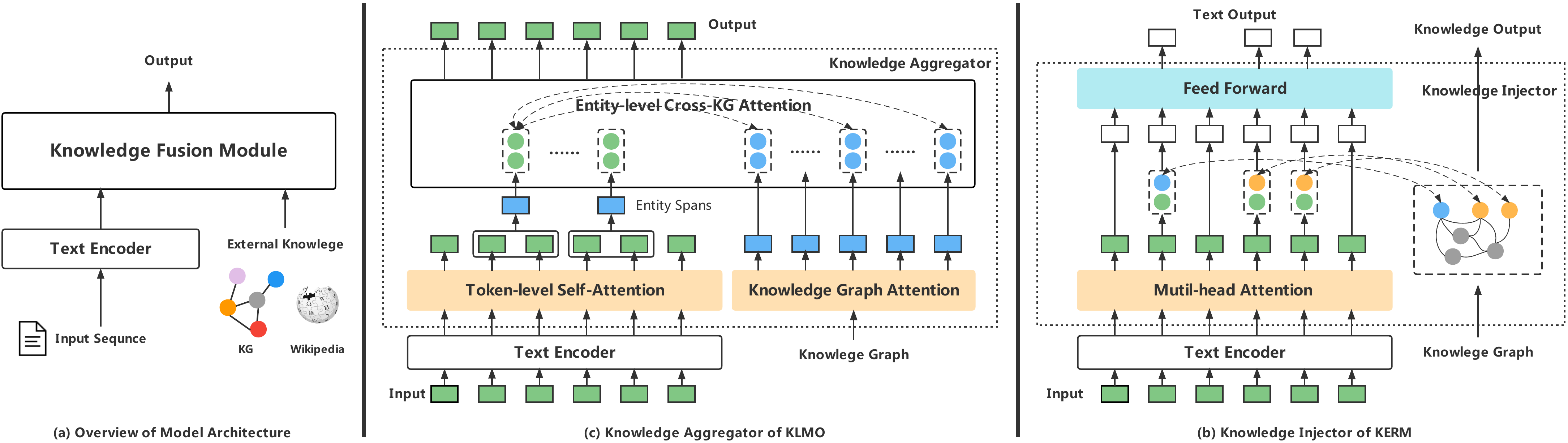}
        \caption{(a) Incorporating triplet knowledge through a knowledge fusion module. (b) KLMO designs a knowledge aggregator to fuse knowledge into the input token sequence. (c) KERM develops a knowledge injector to integrate knowledge explicitly.}
        \label{4}
    \end{figure*}
    
    The third is to change the model structure, which usually introduces a knowledge fusion module \cite{verga2020facts, wang2020k, Emelin2022, he2021klmo, dong2022incorporating, sun2022jointlk, zhang2022greaselm, chen2022knowprompt, ye2022ontology, yu2022jaket}, as shown in Fig. \ref{4}(a).
    FaE \cite{verga2020facts} introduces an additional memory module of facts on the basis of EaE \cite{fevry2020entities}, so that it can effectively combine the information in the symbolic knowledge graph. Besides the layer structure of the original pre-trained model, K-adapter \cite{wang2020k} and KB-adapters \cite{Emelin2022} incorporate knowledge into PLM through external adapter modules. KLMO \cite{he2021klmo} uses a component named knowledge aggregator to fuse the embeddings of the input text and KG, which applies an entity-level cross-KG attention to interactively model entity segments in text along with entities and relations in KG, as shown in \ref{4}(b). KERM \cite{dong2022incorporating} designs a knowledge injector module that combines the knowledge between text corpus and KG for \textcolor{black}{ passage re-ranking task} as shown in Fig. \ref{4}(b). 
    JointLK \cite{sun2022jointlk} and GreaseLM \cite{zhang2022greaselm} exploit  GNNs for modeling extracted knowledge graphs and couple the LM with GNN modules to perform joint reasoning for commonsense reasoning.
     \textcolor{black}{Besides, KnowPrompt \cite{chen2022knowprompt} incorporates entity and relation knowledge contained in the relation labels into prompt templates, and inserts these templates into the input text for relation extraction.} OntoPrompt \cite{ye2022ontology} linearizes ontology knowledge extracted from external knowledge graph into texts as auxiliary prompts, and introduces a \textcolor{black}{visible matrix to guide the knowledge injection process, avoiding injecting irrelevant or noisy knowledge.} 
    
    % Besides, the biggest difference between CoLAKE and K-BERT is that, CoLAKE does not inject triples during fine-tuning, but jointly learns embeddings of entities and relations during pre-training stage. In view of this, they are also representative work under the two categories of explicit incorporation with KG (fusion without changing model's structure), that is, knowledge enhancement during training \cite{sun2020colake} and knowledge enhancement during reasoning \cite{liu2020k}. 
	
    In addition to using knowledge graph as auxiliary information to improve PLMs' ability of language understanding, some work also learns a knowledge embedding representation while incorporating triplet knowledge, so that PLMs can complete some tasks related to knowledge reasoning, such as entity classification, relation prediction, knowledge graph completion and etc. Representative work includes ERICA \cite{qin2020erica}, KEPLER \cite{wang2021kepler}, KLMO \cite{he2021klmo}, FaE \cite{verga2020facts} mentioned above. 
    
    % This has shown that injecting external structured knowledge into the prompt learning process can reduce the cost of template construction and promote the awareness of model on domain tasks.
 
\subsubsection{Fusion stage}
    The above mentioned methods ERNIE \cite{sun2019ernie}, ERNIE-THU\cite{zhang2019ernie}, WKLM\cite{xiong2019pretrained}, LUKE\cite{yamada2020luke}, EaE\cite{fevry2020entities}, FaE\cite{verga2020facts}, ERICA\cite{qin2020erica}, CoLAKE\cite{sun2020colake}, KEPLER\cite{wang2021kepler}, KLMO\cite{he2021klmo}, DKPLM\cite{zhang2022dkplm}, KERM\cite{dong2022incorporating}, JAKET\cite{yu2022jaket} and KP-PLM\cite{wang2022knowledge} are pre-fusion methods that fuse knowledge in the pre-training stage,  while KnowBert\cite{peters2019knowledge}, K-BERT\cite{liu2020k}, K-adapter\cite{wang2020k}, \cite{xu2021human}, KECP\cite{wang2022kecp}, KB-adapters\cite{Emelin2022}, JointLK\cite{sun2022jointlk}, GreaseLM\cite{zhang2022greaselm} are post-fusion methods which fuse knowledge in the fine-tuning stage. 
    The two fusion stages are also called the training stage and reasoning stage, respectively.
    For example, CoLAKE \cite{sun2020colake} jointly learns the embeddings of entities and relations during the  training phase, while K-BERT \cite{liu2020k} injects triples from KG during the reasoning phase.
    %Hence they are also the representative work under two ways of explicit incorporation with KG (fusion without changing model's structure)

\subsection{Incorporating Rule Knowledge into PLMs}
	
	Logic rules always contain clear logical reasoning processes, and can formalize knowledge from external sources \cite{von2019informed}. Incorporating this type of knowledge into PLMs can facilitate the demonstration of reasoning path via its good interpretability. For example, RuleBERT \cite{saeed2021rulebert} utilizes the Horn rules of existing corpus to establish a training dataset and then fine-tunes the model on it. It adopts a probabilistic answer set programming to predict the probability of events, and tries to learn soft rules from PLM. The results show PLMs that reason with soft rules over natural language could improve their performance for deductive reasoning tasks. Besides, PTR \cite{han2021ptr} incorporates logic rules to construct task-specific prompts composed with sub-prompts designed manually, so that the model can encode task-related prior knowledge in the prompt-tuning and generate prompts that are more interpretable. 
    Both RuleBERT and PTR incorporate rule knowledge in the fine-tuning stage.

%-----------------------section 3 KE-PLM in NLG---------------------------------

\section{KE-PLMs for NLG}

    \begin{figure*}[htbp]
        \centering
        \includegraphics[width=\textwidth]{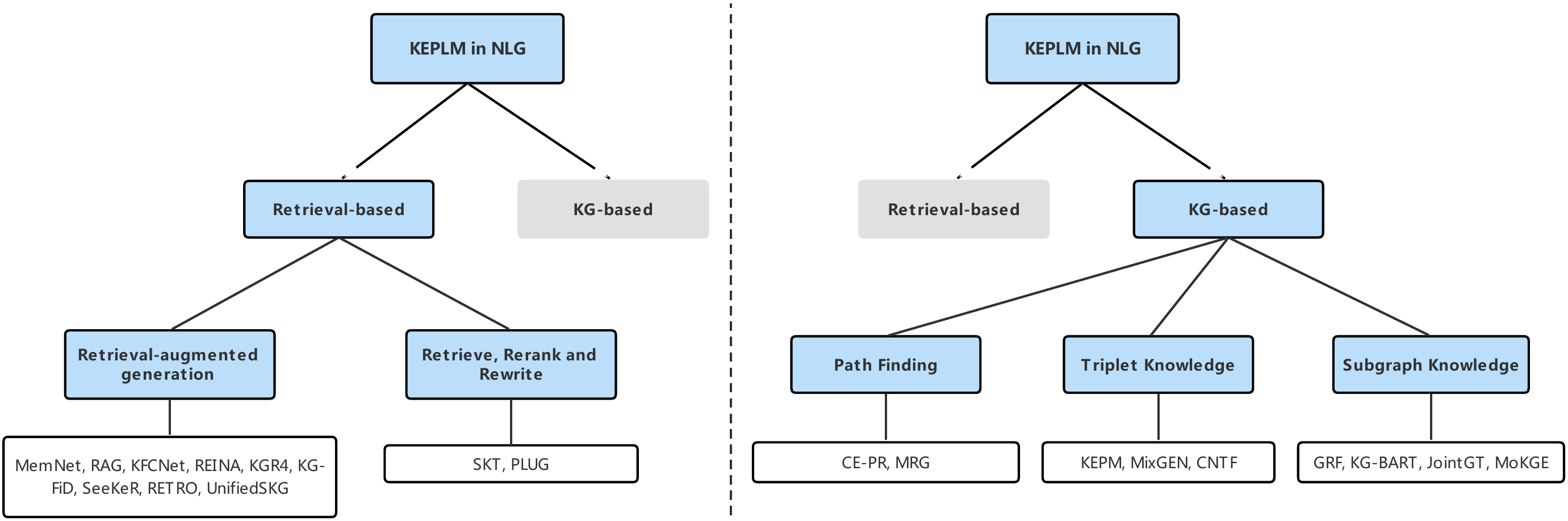}
        \caption{Further categorization of retrieval-based method and KG-based method. The left figure demonstrates the categorization of retrieval-based method, and the right demonstrates the categorization of KG-based method.}
        \label{5}
    \end{figure*}

    The goal of NLG is to enable machines to generate language texts that can be understood by humans and follow the way where humans express themselves. Incorporating various forms of knowledge into generation models other than input sequences helps to improve the performance of text generation tasks. Referring to the survey on knowledge-enhanced text generation \cite{yu2022survey}, we further divide KE-PLMs in the field of NLG into two categories based on their different knowledge sources: one is retrieval-based method and the other is KG-based method. 

\subsection{Incorporating Retrieval-based Knowledge into PLMs}
	
\begin{figure}[!tbp]
        \centering
        \includegraphics[width=\linewidth]{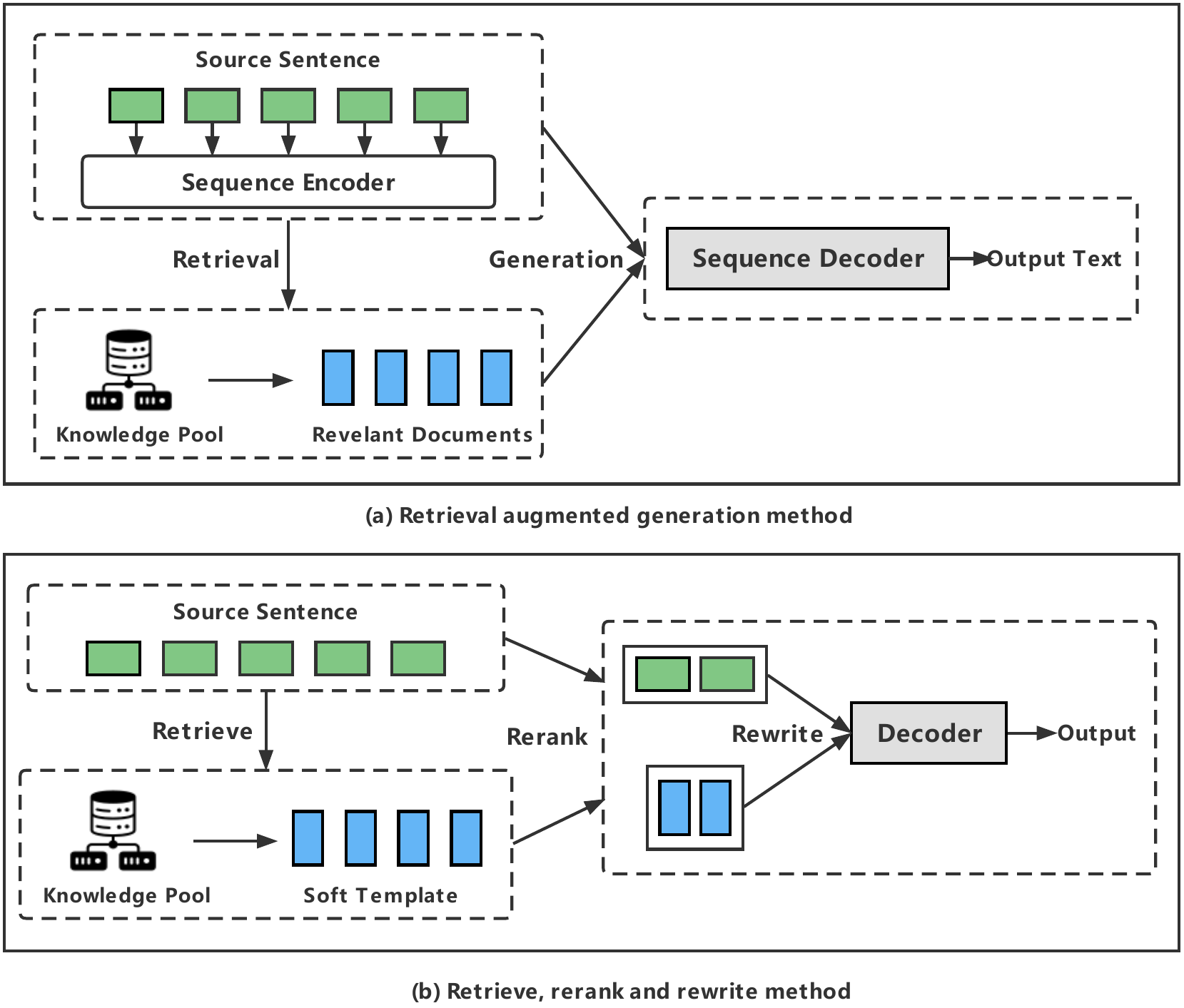}
        \caption{Flow charts of retrieval-based methods. (a) demonstrates the retrieval augmented generation method. (b) demonstrates the retrieve, rerank and rewrite method. Notably, the retrieved candidate templates will be reranked  before rewriting.}
        \label{6}
    \end{figure}

	The retrieval-based methods mainly integrate and utilize additional knowledge related to the input sequences through retrieval. Other than the input sequence itself, the additional knowledge is retrieved from external sources such as online search engines, large data sets, and training sets to guide the  generation process.
 %retrieved items will be further selected/edited before being used for generation, 
    \textcolor{black}{Considering whether the methods focus on re-ranking the retrieved items for generation or not, we divide these retrieval-based methods into two sub-methods as shown in the left of Fig. \ref{5}: one is retrieval augmented generation method that aims to improve generation by retrieving related knowledge  \cite{dinan2018wizard, lewis2020retrieval, li2021kfcnet, wang2022training, liu2022kgr4, yu2021kg, shuster2022language, borgeaud2022improving, Xie2022UNIFIEDSKG}, and the other is retrieve, rerank and rewrite method which focuses on re-ranking retrieved items for generation \cite{kim2020sequential, Li2022PLUG}.}
    \textcolor{black}{The flow charts of these two kinds of methods are shown in Fig. \ref{6}}. Notably, these KE-PLMs incorporate external knowledge in the fine-tuning stage to improve their performance on downstream tasks.
    % Based on the different  retrieval stages \cite{yu2022survey}, we  divide them into two sub-methods under this category as shown in the left of Figure 5: one is retrieval augmented generation method which contains retrieval and generation these two stages \cite{dinan2018wizard, lewis2020retrieval, li2021kfcnet, wang2022training, liu2022kgr4, yu2021kg, shuster2022language, borgeaud2022improving}, the other is retrieve, rerank and rewrite method which contains three stages \cite{kim2020sequential}. 
	
    % scale=0.125
	
    % MemNet \cite{dinan2018wizard}, RAG \cite{lewis2020retrieval}, KFCNet \cite{li2021kfcnet}, all integrate external knowledge into NLG tasks by introducing retriever components.

% decoder contrastive module enhances the utility of retrieved prototypes while learning general features.
% The core idea of contrastive learning is to construct positive and negative samples from an anchor sample, and draw together the anchor and positive samples while pushing away the anchor from all negative samples in the embedding space during training. Given that high-quality prototypes can be used as clusters of positive samples, we propose a decoder contrastive module that minimizes the distance between decoded sentence representations with distinct prototypes retrieved from the same concept set. Common scenario information and abstract concept relationships can be learned based on the contrasts between different prototypes. 

    In the line of retrieval augmented generation method, MemNet \cite{dinan2018wizard} proposes a Transformer memory network that can retrieve topic-relevant knowledge based on the dialogue history, and then generates the next dialogue utterance with the help of this retrieved knowledge.
    RAG \cite{lewis2020retrieval} leverages a retriever to find the top-K relevant documents given the input sequence, and uses them as additional context when generating predictions.
    \textcolor{black}{KFCNet \cite{li2021kfcnet} first retrieves prototypes that comprise concepts in the given concept set, while keeping these retrieved results semantically similar to the target sentences. Then it applies two contrastive learning modules in both encoder and decoder to capture global target information and learn general features from the multiple retrieved prototypes.} 
    REINA \cite{wang2022training} retrieves some training samples which are similar to the input text and takes them as knowledge to improve the effect of machine translation. KGR4 \cite{liu2022kgr4} divides the generation of commonsense into four stages, that is, retrieval of commonsense, generation of commonsense by means of generation models, refinement and correction of the generated commonsense statements, and scoring of the generated statements. Best results can be obtained through this process. KG-FiD \cite{yu2021kg} is based on the Fusion-in-Decoder (FID) \cite{izacard2020leveraging} model, but it proposes to solve the validity and efficiency of FID with the help of KG, which improves the ability of open domain question answering significantly. SeeKeR \cite{shuster2022language} leverages a set of documents retrieved from the search engine to generate knowledge response. This method can incorporate up-to-date information through its three-module framework (search, knowledge generation, and final response). RETRO \cite{borgeaud2022improving} retrieves from large text databases and builds trillions of tokens as retrieve sources to expand the language model. 
    \textcolor{black}{UnifiedSKG \cite{Xie2022UNIFIEDSKG} unifies six categories of tasks  (i.e., semantic parsing, question answering, data-to-text generation, fact verification, conversation, and formal language-to-text translation tasks) into a text-to-text format and introduces linearized knowledge to strengthen their performance.}
	
	Representative work of the retrieve, rerank and rewrite method, such as SKT \cite{kim2020sequential}, regards knowledge selection as a sequential decision-making process, and uses the sequential latent variable model to improve the accuracy of knowledge selection in multiple rounds of dialogue. 
    % 新添加的工作
    \textcolor{black}{PLUG \cite{Li2022PLUG} retrieves related knowledge from Wikipedia, dictionary, and knowledge graph, and then ranks them based on statistical and semantic information for knowledge-grounded dialogue generation.}
	
\subsection{Incorporating KG-based Knowledge into PLMs}
	
	In order to distinguish the granularity of knowledge utilized by different KE-PLMs more specifically, we divide existing work into three categories: knowledge extracted by path finding, triplet knowledge, and subgraph knowledge extracted from KG as shown in the right of Fig. \ref{5}.  
	
	The first is to extract knowledge by path finding \cite{ji2020generating, zhao2020graph}. In this way, the relation path is clearly reasoned to make reliable decisions. CE-PR \cite{ji2020generating} and MRG \cite{zhao2020graph} mainly perform explicit reasoning on relation paths,  significantly improving the effectiveness of text generation. Specifically, 
    \textcolor{black}{CE-PR first retrieves a subgraph given the source concepts, and then scores each triple on this graph, propagates scores along the paths to each node from the source concepts and preserves the nodes with higher scores.}
   % and finally select the ones bases on their selection probability
    MRG leverages the reasoning module to infer paths step by step from the source concepts, and then uses the sentence realization (i.e., sentence generation) module to generate a complete sentence based on the inferred paths. Fig. 7 demonstrates the inference process of these two methods.

    % CE-PR first retrieves a subgraph from the source concepts, and then scores each triple in the subgraph, routes paths along the connected nodes, and selects concepts from them.
	
    \begin{figure}[!tbp]
        \centering
        \includegraphics[width=0.9\linewidth]{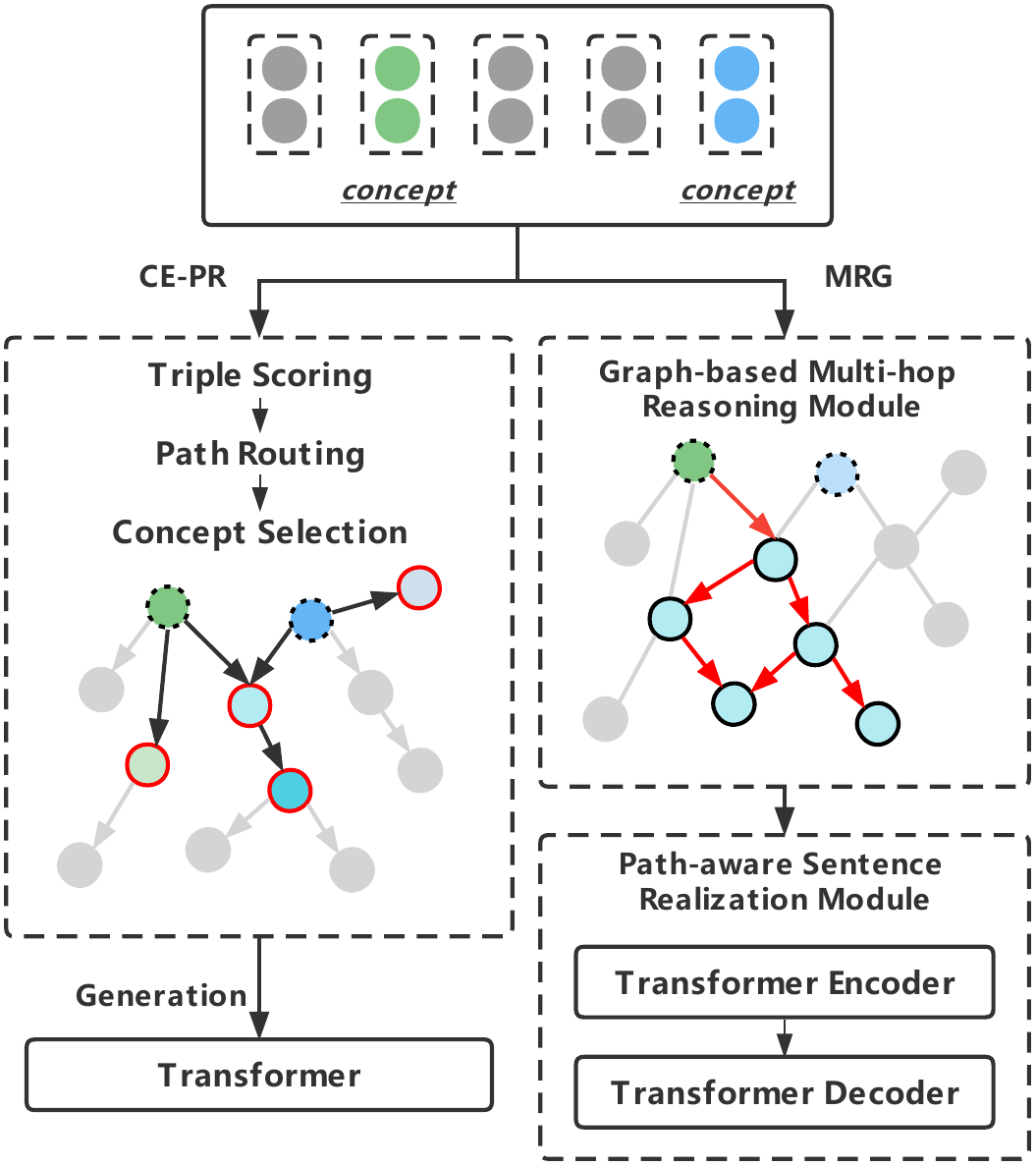}
        \caption{Comparison of CE-PR and MRG incorporating knowledge extracted by path finding. }
        \label{7}
    \end{figure}
	
	The second sub-category is KE-PLMs based on triplet knowledge \cite{guan2020knowledge, Sridhar2022MinGEN, Varshney2022CNTF}. KEPM \cite{guan2020knowledge} \textcolor{black}{converts commonsense triplets in the knowledge bases into natural language statements based on templates  to provide additional information for story generation}. 
    \textcolor{black}{MixGEN \cite{Sridhar2022MinGEN} adopts an encoder-decoder framework to incorporate  expert knowledge (from dataset annotation),  explicit knowledge (from knowledge graph), and implicit knowledge (from generative PLM) to produce implications of toxic text.} \textcolor{black}{CNTF \cite{Varshney2022CNTF} models the commonsense, named entities and topic-specific knowledge via a multi-hop attention module to facilitate dialogue generation task.}

	The third one is subgraph knowledge \cite{ji2020language, liu2021kg, ke2021jointgt, yu2022diversifying}. Different from the first two categories, subgraph contains context information on related concepts, which plays an important role in understanding related concepts and language generation. This kind of methods generally use GNNs to model extracted subgraphs, and then fuse these extracted subgraphs to enhance the natural language generation ability.
	
\begin{figure}[htbp]
        \centering
        \includegraphics[width=0.8\linewidth]{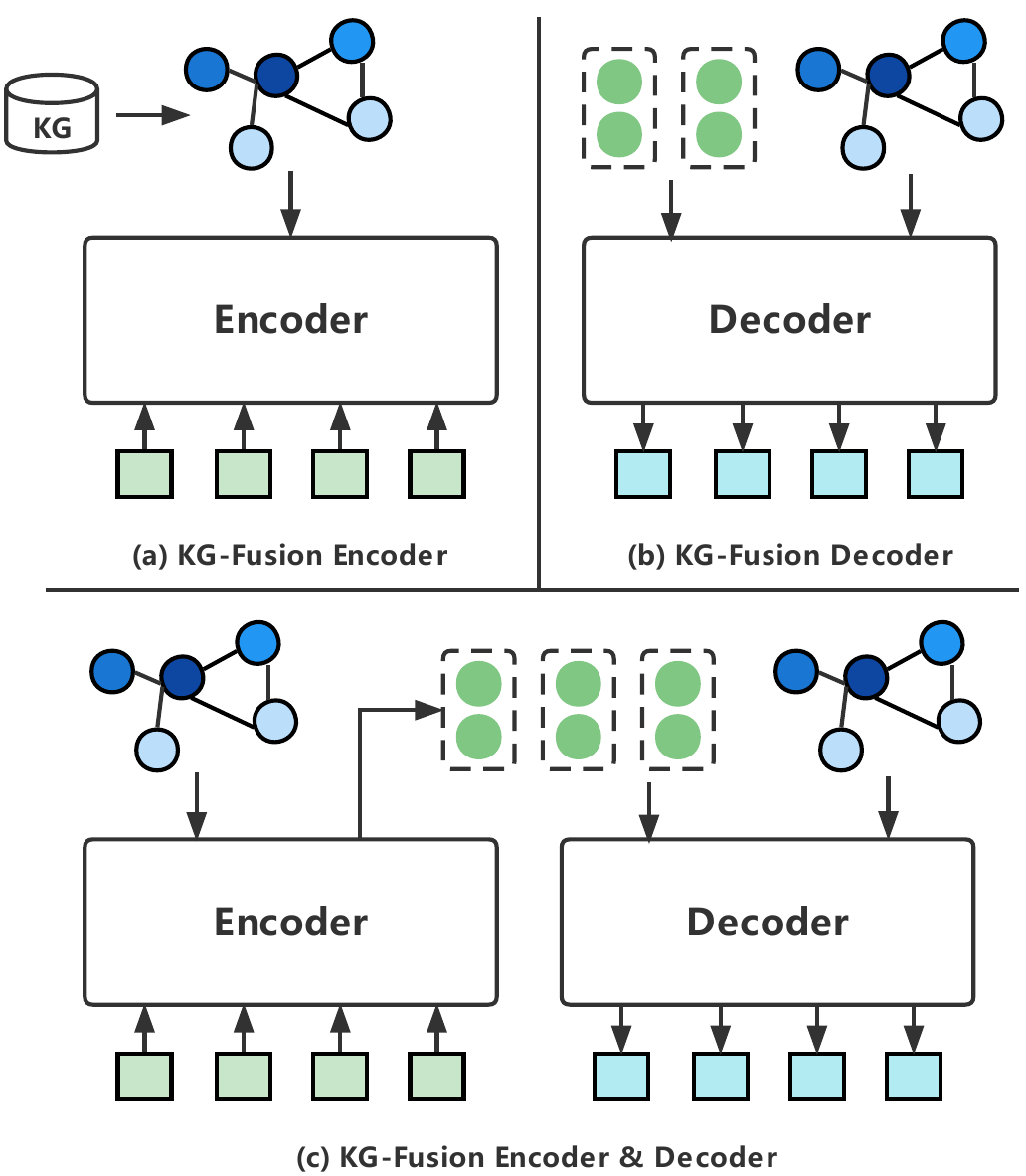}
        \caption{\textcolor{black}{Integration position of subgraph knowledge. (a)  incorporating KG into the encoder. (b) incorporating KG into the decoder. (c) incorporating KG into both the encoder and decoder.}}
        \label{8}
    \end{figure}

	For the integration of subgraph knowledge, we further divide them based on the specific position they integrate as shown in Fig. \ref{8}. Some work incorporates subgraph knowledge into only encoder to improve the language understanding ability, such as JointGT \cite{ke2021jointgt} and MoKGE\cite{yu2022diversifying}. Some inject subgraph knowledge into the decoder, such as GRF \cite{ji2020language}. In this way, every step of decoding can be traced, so that the model can be better interpreted. Others introduce subgraph knowledge into both encoder and decoder, such as KG-BART \cite{liu2021kg}. 
	
    Among the above mentioned KE-PLMs for NLG, JointGT\cite{ke2021jointgt}, PLUG\cite{Li2022PLUG}, UnifiedSKG\cite{Xie2022UNIFIEDSKG} are pre-fusion methods (fuse knowledge during pre-training),  and CE-PR\cite{ji2020generating}, MRG\cite{zhao2020graph}, KEPM\cite{guan2020knowledge}, GRF\cite{ji2020language}, KG-BART\cite{liu2021kg}, MoKGE\cite{yu2022diversifying}, MixGEN\cite{Sridhar2022MinGEN}, CNTF\cite{Varshney2022CNTF} are post-fusion methods (fuse knowledge during fine-tuning).

    In Table \ref{10} and Table \ref{11}, we respectively summarize the existing KE-PLMs for NLU and NLG.
    
    % \onecolumn
    \begin{table*}[]
		\caption{Summarization of different KEPLMs. Here RC: relation classification; ET: entity typing; QA: question answering; RE: relation extraction; WiC: words in context; NER: named entity recognition; NLI: natural language inference; GLUE: general language understanding evaluation.}
		\label{10}
		\setcellgapes{1.5pt}
		\makegapedcells
		\resizebox{\textwidth}{!}{%
			\begin{tabular}{l|l|l|l|l}
				\hline
				\makecell[l]{Method} & Knowledge Type & \makecell[l]{Fusion in \\Pre-training} & \makecell[l]{Fusion in \\Fine-tuning} & NLU or NLG tasks \\ 
				\hline
				LIBERT \cite{lauscher2019specializing} & lexical & Yes & & \makecell[l]{lexical simplification, sentence/sentence-pair \\classification, NLI} \\
				\hline
				LIMIT-BERT \cite{zhou2019limit} & syntax tree & Yes & & syntactic parsing, semantic parsing \\
				\hline
				SenseBERT \cite{levine2019sensebert} & lexical & Yes & & word supersense disambiguation, WiC \\
				\hline
				SKEP \cite{tian2020skep} & lexical & Yes & & \makecell[l]{sentence/aspect-level sentiment \\classification, opinion role labeling}\\
				\hline
				Sentiprompt \cite{li2021sentiprompt} & lexical &  & Yes & \makecell[l]{triplet extraction, pair extraction \\aspect term extraction}\\
				\hline
				LET \cite{lyu2021let} & lexical & & Yes & Chinese short text matching \\
				\hline
				KEAR \cite{xu2021human} & lexical, general text, triplet & & Yes & commonsense reasoning \\
				\hline
				Syntax-BERT \cite{bai2021syntax} & syntax tree & Yes & & \makecell[l]{syntactic and semantic compositionality of \\sentiment classification, NLI, GLUE}\\
				\hline
				DictBERT \cite{chen2022dictbert} & lexical & Yes & & NER, RE, commonsense reasoning \\
				\hline
				
				KNN-LM \cite{khandelwal2019generalization} & general text & & Yes & language modeling \\
				\hline
				REALM \cite{guu2020retrieval} & general text & Yes & & open-QA \\
				\hline
				ExpBERT \cite{murty2020expbert} & general text & & Yes & RE\\
				\hline
                OK-Transformer \cite{cui2022enhancing} & general text & & Yes & commonsense reasoning, text classification \\
                \hline
				Kformer \cite{yao2022kformer} & general text & & Yes & commonsense reasoning, medical QA \\
				\hline
				REINA \cite{wang2022training} & \makecell[l]{general text, retrieval \\augmented generation} & & Yes &  \makecell[l]{summarization, language modeling \\machine translation, QA} \\
				\hline
				UniK-QA \cite{oguz2020unik} & general text & & Yes & multi-source QA \\
				\hline
				UDT-QA \cite{ma2022open} & general text & & Yes & open-domain QA \\
				\hline
				BioBERT \cite{lee2020biobert} & domain-specific text & Yes & & biomedical NER, RE, QA\\
				\hline
				SciBERT \cite{beltagy2019scibert} & domain-specific text & Yes & & \makecell[l]{sequence tagging, sentence classification, \\dependency parsing}\\
				\hline
				S2ORC-BERT \cite{lo2019s2orc} & domain-specific text & Yes & & \makecell[l]{inline citation detection, bibliography \\parsing, bibliography linking} \\
				\hline
		
				ERNIE \cite{sun2019ernie} & entity & Yes & & \makecell[l]{NLI, semantic similarity, NER, \\sentiment analysis, QA} \\
				\hline
				ERNIE-THU \cite{zhang2019ernie} & entity & Yes & Yes & ET, RC \\
				\hline
				WKLM \cite{xiong2019pretrained} & entity & Yes & & QA, ET \\
				\hline
				KnowBert \cite{peters2019knowledge} & entity & Yes & Yes & RE, WiC, ET \\
				\hline
				LUKE \cite{yamada2020luke} & entity & Yes & & ET, RC, NER, cloze-style/extractive QA \\
				\hline
				EaE \cite{fevry2020entities} & entity & Yes & & open-domain QA, RE \\
				\hline
				KECP \cite{wang2022kecp} & entity & & Yes & extractive QA  \\
				\hline
				
				ERICA \cite{qin2020erica} & entity, triplet & Yes & & RE, ET, QA \\
				\hline
				K-BERT \cite{liu2020k} & triplet & & Yes & sentence classification, QA, NER \\
				\hline
				CoLAKE \cite{sun2020colake} & triplet & Yes & & \makecell[l]{ET, RE, knowledge probing, GLUE \\knowledge graph completion} \\
				\hline
				FaE \cite{verga2020facts} & entity, triplet & Yes & & open-domain QA \\
				\hline
				K-adapter \cite{wang2020k} & general text, triplet & & Yes & ET, QA, RC \\
				\hline
                \textcolor{black}{KB-adapters} \cite{Emelin2022} & entity,  triplet & & Yes & \makecell[l]{knowledge-probing using response selection \\fact memorization, response generation}  \\
				\hline
				KLMO \cite{he2021klmo} & entity, triplet & Yes & & ET, RC \\
				\hline
				KEPLER \cite{wang2021kepler} & entity, triplet & Yes & & RC, ET, GLUE, link prediction \\
				\hline
				DKPLM \cite{zhang2022dkplm} & entity, triplet & Yes & & knowledge probing, RE, ET \\
				\hline
                JAKET\cite{yu2022jaket} & entity, triplet & Yes & & RC, QA over KG, entity classification \\
                \hline
                KP-PLM \cite{wang2022knowledge} & triplet & Yes & & knowledge probing, RE, ET \\
				\hline
				KERM \cite{dong2022incorporating} & triplet & Yes & & passage re-ranking \\
				\hline
                JointLK \cite{sun2022jointlk} & triplet & & Yes & commonsense reasoning \\
				\hline
                GreaseLM \cite{zhang2022greaselm} & triplet & & Yes & commonsense reasoning \\
				\hline 
				RuleBERT \cite{saeed2021rulebert} & rule & & Yes & rule reasoning \\
				\hline
				PTR \cite{han2021ptr} & rule & & Yes & RC \\
				\hline
			\end{tabular}
		}
	\end{table*}	
    % \twocolumn
    
    \begin{table*}[]
		\caption{Summarization of different KEPLMs. Here QA: question answering.}
		\label{11}
		\setcellgapes{1.5pt}
		\makegapedcells
		\resizebox{\textwidth}{!}{%
			\begin{tabular}{l|l|l|l|l}
				\hline
				\makecell[l]{Method} & Knowledge Source & \makecell[l]{Fusion in \\Pre-training} & \makecell[l]{Fusion in \\Fine-tuning} & NLU or NLG tasks \\
				\hline
				MemNet \cite{dinan2018wizard} & retrieved text & & Yes & open-domain dialogue generation \\
				\hline
				RAG \cite{lewis2020retrieval} & retrieved text & & Yes & \makecell[l]{open-domain/abstractive QA, \\question generation, fact verification} \\
				\hline
				SKT \cite{kim2020sequential} & retrieved text & & Yes & knowledge-grounded dialogue  \\
				\hline
				KFCNet \cite{li2021kfcnet} & retrieved text & & Yes & \makecell[l]{commonsense/keyword generation}\\
				\hline
				KG-FiD \cite{yu2021kg} & retrieved text & & Yes & open-domain QA \\
				\hline
				KGR4 \cite{liu2022kgr4} & retrieved text & & Yes & commonsense generation \\
				\hline
				SeeKeR \cite{shuster2022language} & retrieved text & & Yes & open-domain dialogue, prompt completion \\
				\hline
				RETRO \cite{borgeaud2022improving} & retrieved text & & Yes & language modelling, QA \\
				\hline
                \textcolor{black}{UnifiedSKG} \cite{Xie2022UNIFIEDSKG} & structured knowledge & Yes & & structured knowledge grounding \\
				\hline
                \textcolor{black}{PLUG} \cite{Li2022PLUG} & \makecell[l]{retrieved text,\\ knowledge graph} & Yes &  & dialogue generation \\
				\hline
				
				CE-PR \cite{ji2020generating} & knowledge graph & & Yes & commonsense explanation generation \\
				\hline
				MRG \cite{zhao2020graph} & knowledge graph & & Yes & story/review/description generation\\
				\hline
				KEPM \cite{guan2020knowledge} & knowledge graph & & Yes & story generation\\
				\hline
                \textcolor{black}{MixGEN} \cite{Sridhar2022MinGEN} & \makecell[l]{knowledge graph,\\expert knowledge} & & Yes & toxic text explanation\\
				\hline
                \textcolor{black}{CNTF} \cite{Varshney2022CNTF} & knowledge graph & & Yes & dialogue generation\\
				\hline
				GRF \cite{ji2020language} & knowledge graph & & Yes & \makecell[l]{story ending generation, abductive NLG, \\explanation generation} \\
				\hline
				KG-BART \cite{liu2021kg} & knowledge graph & & Yes & commonsense  generation/QA \\
				\hline
				JointGT \cite{ke2021jointgt} & knowledge graph & Yes & & KG-to-text generation \\
				\hline
				MoKGE\cite{yu2022diversifying} & knowledge graph & & Yes & \makecell[l]{commonsense explanation generation, \\abductive commonsense reasoning} \\
				\hline
				
			\end{tabular}
		}
	\end{table*}

%-----------------------section 4 Future---------------------------------

\section{FUTURE DIRECTIONS}

    In this section, we propose some possible research directions of KE-PLMs in the future, which may meet the existing problems and challenges.

\subsection{Integrating Knowledge from Homogeneous and Heterogeneous Sources}

    Since most of the existing work only utilizes knowledge from a single source, such as knowledge graph or web resource, exploring how to integrate knowledge from heterogeneous sources is still a valuable direction for future research. 
	
    As we present in the section above, some prior work has tried to incorporate different types of knowledge to improve the performance of question-answering. For example, UniK-QA \cite{oguz2020unik} integrates external knowledge including text, tables, and relational triplets in the knowledge base. Through the heuristic method of linearizing heterogeneous knowledge sources including knowledge base (KB) triples and semi-structured tables into text, it unifies structured knowledge involved in KBQA and unstructured knowledge involved in TextQA, expanding the sources of external knowledge. UDT-QA \cite{ma2022open} introduces structured knowledge such as knowledge graphs and tables into open-domain question answering, and converts them into linear sequences as the input of text generation tasks. 
	
    In the field of open-domain question answering, improving the ability of PLMs to integrate multiple knowledge sources can effectively increase the knowledge coverage, so that models can generate more reliable answers.

\subsection{Exploring Multi-modal Knowledge}
	
    Most of the current research focuses merely on text knowledge with fewer multi-modal sources. In fact, images, videos, and audio in addition to textual and tabulated information can also become the knowledge sources of PLMs, which can further improve the performance of KE-PLMs. 
	
    Several studies have explored integrating multi-modal knowledge. Representative work includes KB-VLP \cite{chen2021kb-vlp}, and ERNIE-VIL \cite{yu2021ernie}. 
    \textcolor{black}{KB-VLP \cite{chen2021kb-vlp} extracts knowledge information from the external knowledge base based on both the input text and image, and uses the knowledge as additional inputs to enhance the model's ability of semantic alignment and knowledge perception.}
    \textcolor{black}{ERNIE-VIL \cite{yu2021ernie} parses the input description texts of images into structured scene graphs, and designs cross-modal pre-training tasks to pay attention to detailed semantic alignments across vision and language modalities.}
	
    Since images and associated text contain rich semantics, the injection of these different modalities of knowledge and concentration on detail semantics can make them complement and enhance each other, which will boost the performance of PLMs on both NLU and NLG tasks.

    % KB-VLP focuses on three modalities of information including text, image and  knowledge graph. 
    % \textcolor{red}{By extracting knowledge graph embeddings from text and detected image tags, it makes full use of commonsense knowledge and logical reasoning, which improves the model's ability of semantic alignment and knowledge perception.} 
    % parses various elements by constructing scene graphs from text, which better represents the fine-grained semantic information required between different modalities.

\subsection{Providing Interpretability Evidence}
	
    Although many existing KE-PLMs have achieved great success on a series of text generation tasks, it should not be ignored that, if the generation process requires commonsense knowledge reasoning, the performance of models will be affected.
	
    Some work has attempted to tackle this problem \cite{lewis2020retrieval, guu2020retrieval, wang2020k}. For example, GRF \cite{ji2020language} utilizes external knowledge graphs for explicit commonsense reasoning, and incorporates rich structural information  in order to perform dynamic multi-hop reasoning on multiple relational paths. Reasoning paths obtained in this process provide a theoretical basis for the generation of results. This work suggests that, giving an explicit reasoning path will help improve the interpretability of models and make predictions more rational.

\subsection{Learning Knowledge in a Continuous Way}
	
    Existing work is usually trained on a large number of static or non-updated data in the pre-training stage. But models may forget the original knowledge learned before when facing new tasks, which leaves them vulnerable to a phenomenon called catastrophic forgetting problem \cite{van2019three}. With the continuous growth of knowledge from heterogeneous sources, exploring methods to make models master new knowledge while not forgetting the previous one learned in the past requires continual learning (also called life-long learning) to integrate various knowledge constantly.
	
    ELLE \cite{qin2022elle} proposes an extension module that maintains the network function to expand the width and depth of the model, so that the model can effectively acquire new knowledge and retain the old to a greater extent at the same time. K-adapter \cite{wang2020k} and \textcolor{black}{KB-adapters} \cite{Emelin2022} adds the adapters with PLMs to store factual and linguistic knowledge, so as to continuously incorporate more knowledge into PLM. 
	
    Incorporating knowledge continuously is a promising direction in future research \cite{yu2022survey}. The application of continuous and increasing pre-training will effectively improve the universality of PLMs, and solve the catastrophic forgetting problem while incorporating more knowledge.

\subsection{Optimizing the Efficiency of Incorporating Knowledge into Large Models}
	
    The scale of pre-trained models and injection of knowledge has become increasingly large in recent years \cite{han2021pre}, thus bringing severe challenges to the computational efficiency and computational resources that cannot be ignored. Though most of the existing work has achieved good results in various pre-training tasks, few studies mention the cost of knowledge fusion in the process.
	
    In view of this challenge, we propose the following two possible directions that may be worth further exploring: one is to improve the efficiency of knowledge acquisition and filtering, and the other is to optimize the computational burden.
	   
    Existing work, such as ZeRO \cite{rajbhandari2020zero}, has been explored in the second area. Based on traditional data parallel training mode, ZeRO deeply optimizes the redundant space and eliminates the memory occupied by redundancy through dividing the parameters, gradients, and optimizer states of the model into different processes.

\subsection{Increasing the Variety of Results Generated}
	
    It is a vital research direction in NLG to generate alternative outputs or predict all possible results for the real situation, which is also the purpose of output diversity in the generative commonsense reasoning task. Existing work, such as MoKGE \cite{yu2022diversifying}, uses the diversified knowledge reasoning of commonsense knowledge graph to complete the diversity generation of NLG. Based on the observation of human annotations, the concepts related to original input are associated into the generation process, and the mixture of expert method is used to generate diversified reasonable outputs, thus increasing the diversity of generated results.

%-----------------------section 5 conclusion---------------------------------

\section{Conclusion}
    
    In this survey, we present a comprehensive review of KE-PLMs from the perspective of NLU and NLG, and respectively propose proper taxonomies for both NLU and NLG to highlight their different focuses. We also discuss the representative work in the taxonomies. Finally, in view of the existing problems and challenges, we discuss potential future research directions of KE-PLMs, hoping to facilitate relevant research in this promising area.

% use section* for acknowledgment
% \ifCLASSOPTIONcompsoc
%   % The Computer Society usually uses the plural form
%   \section*{Acknowledgments}
% \else
%   % regular IEEE prefers the singular form
%  % \section*{Acknowledgment}
% \fi

%The authors would like to thank...

% Can use something like this to put references on a page
% by themselves when using endfloat and the captionsoff option.
\ifCLASSOPTIONcaptionsoff
  \newpage
\fi

\normalem
\bibliographystyle{IEEEtran}
\bibliography{keplm-manuscript}

% \begin{thebibliography}{101}

% \end{thebibliography}

% 
% If you have an EPS/PDF photo (graphicx package needed) extra braces are
% needed around the contents of the optional argument to biography to prevent
% the LaTeX parser from getting confused when it sees the complicated
% \includegraphics command within an optional argument. (You could create
% your own custom macro containing the \includegraphics command to make things
% simpler here.)
%\begin{IEEEbiography}[{\includegraphics[width=1in,height=1.25in,clip,keepaspectratio]{mshell}}]{Michael Shell}
% or if you just want to reserve a space for a photo:

\begin{IEEEbiography}[{\includegraphics[width=1in,height=1.25in,clip,keepaspectratio]{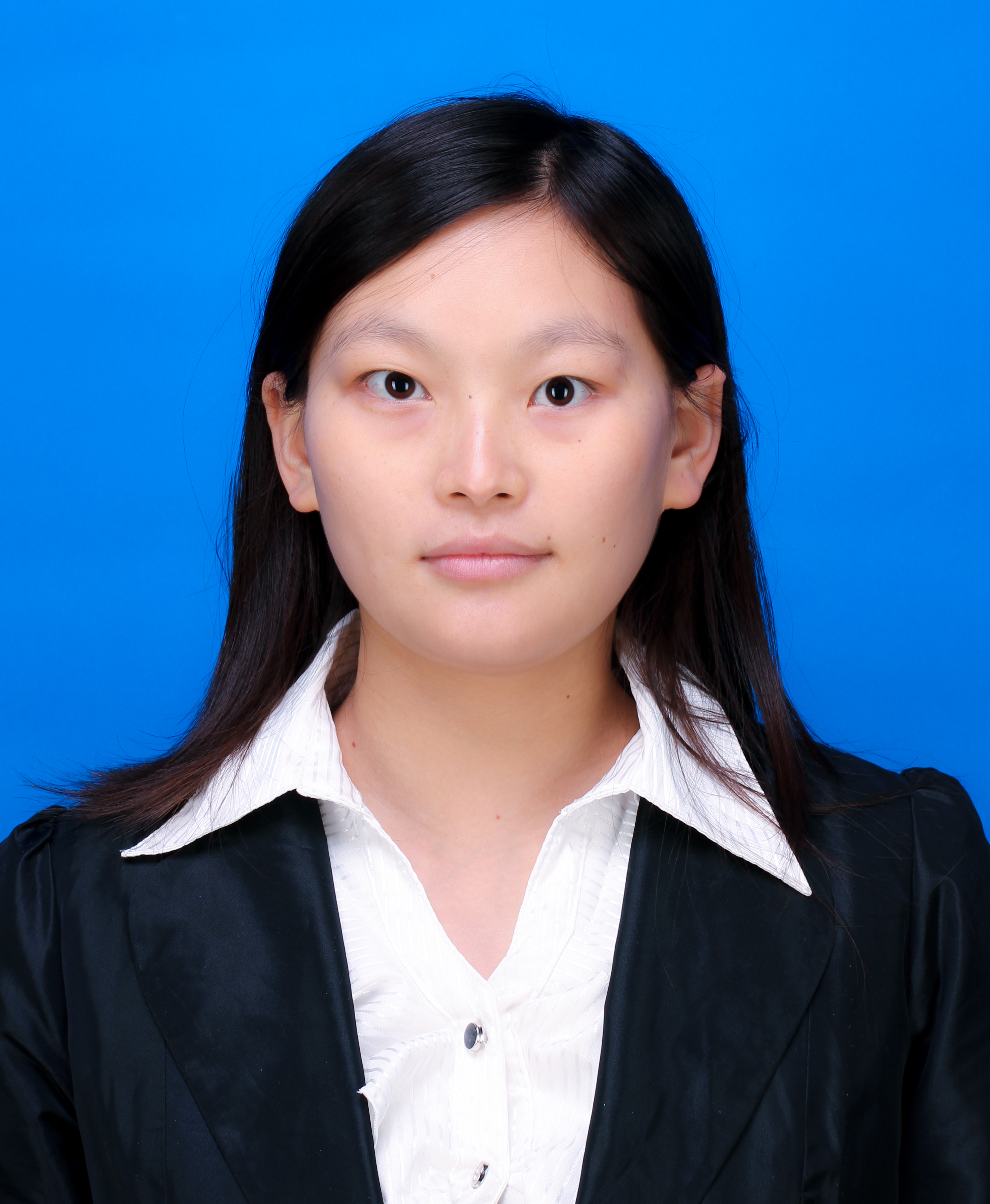}}]{Linmei Hu}
Linmei Hu works as an associate professor in the School of Computer Science and Technology, Beijing Institute of Technology.  She received her Ph.D degree in Tsinghua University in 2018. Her research interests include  Knowledge Graph, Natural Language Processing and Multimodal Content Analysis. She has published some papers in top conferences and journals, such as ACL, AAAI, EMNLP, ACM
SIGKDD, MM, TOIS, and IEEE TKDE.
\end{IEEEbiography}

% if you will not have a photo at all:
\begin{IEEEbiography}[{\includegraphics[width=1in,height=1.25in,clip,keepaspectratio]{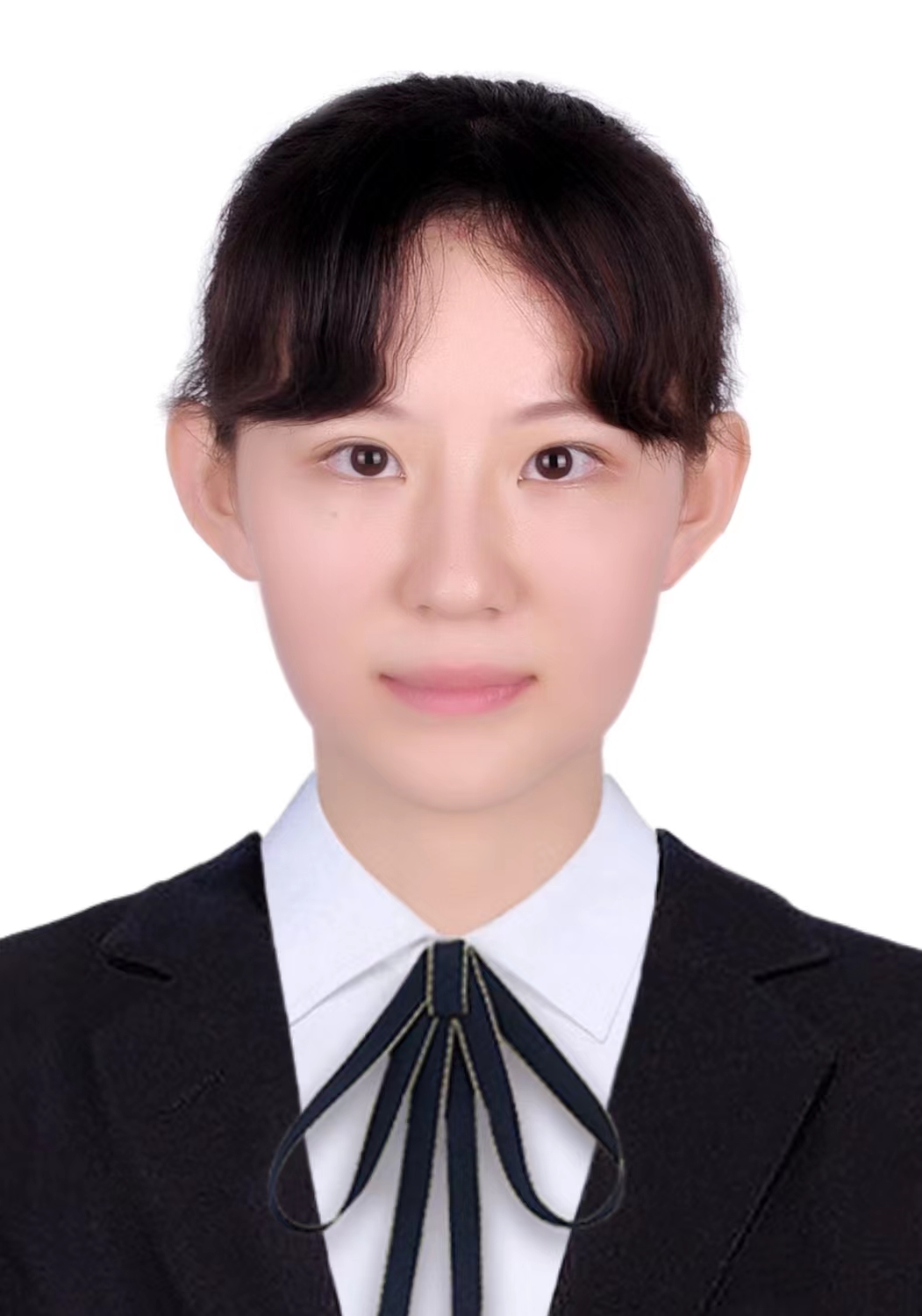}}]
{Zeyi Liu}
Zeyi Liu will receive the B.S. degree from University of Science and Technology Beijing in 2023, and continue to pursue her master's Degree in computer science and technology at Beijing University of Posts and Telecommunications.
\end{IEEEbiography}

\begin{IEEEbiography}[{\includegraphics[width=1in,height=1.25in,clip,keepaspectratio]{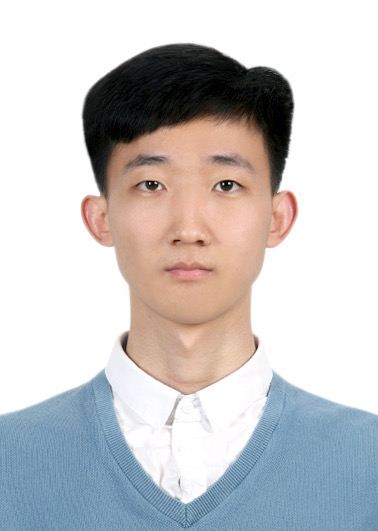}}]{Ziwang Zhao}
Ziwang Zhao received the B.S. degree from Xidian University, Xi'an, China, in 2021. He is currently pursuing the Master's Degree in computer science and technology at Beijing University of Posts and Telecommunications. His research interests include Pre-Trained Models, Knowledge Graph and Multimodal.
\end{IEEEbiography}

\begin{IEEEbiography}[{\includegraphics[width=1in,height=1.25in,clip,keepaspectratio]{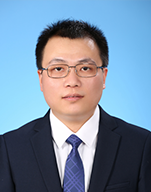}}]{Lei Hou} 
Lei Hou is an assistant professoer in Department of Computer Science and Technology, Tsinghua University. He obtained his Ph.D. degree from Tsinghua University in 2016. His research interests include knowledge graph construction and application, news and user-generated content mining. He has published many research papers on top-tier international conferences, such as ACL, EMNLP, AAAI, etc. 
\end{IEEEbiography}

\begin{IEEEbiography}[{\includegraphics[width=1in,height=1.25in,clip,keepaspectratio]{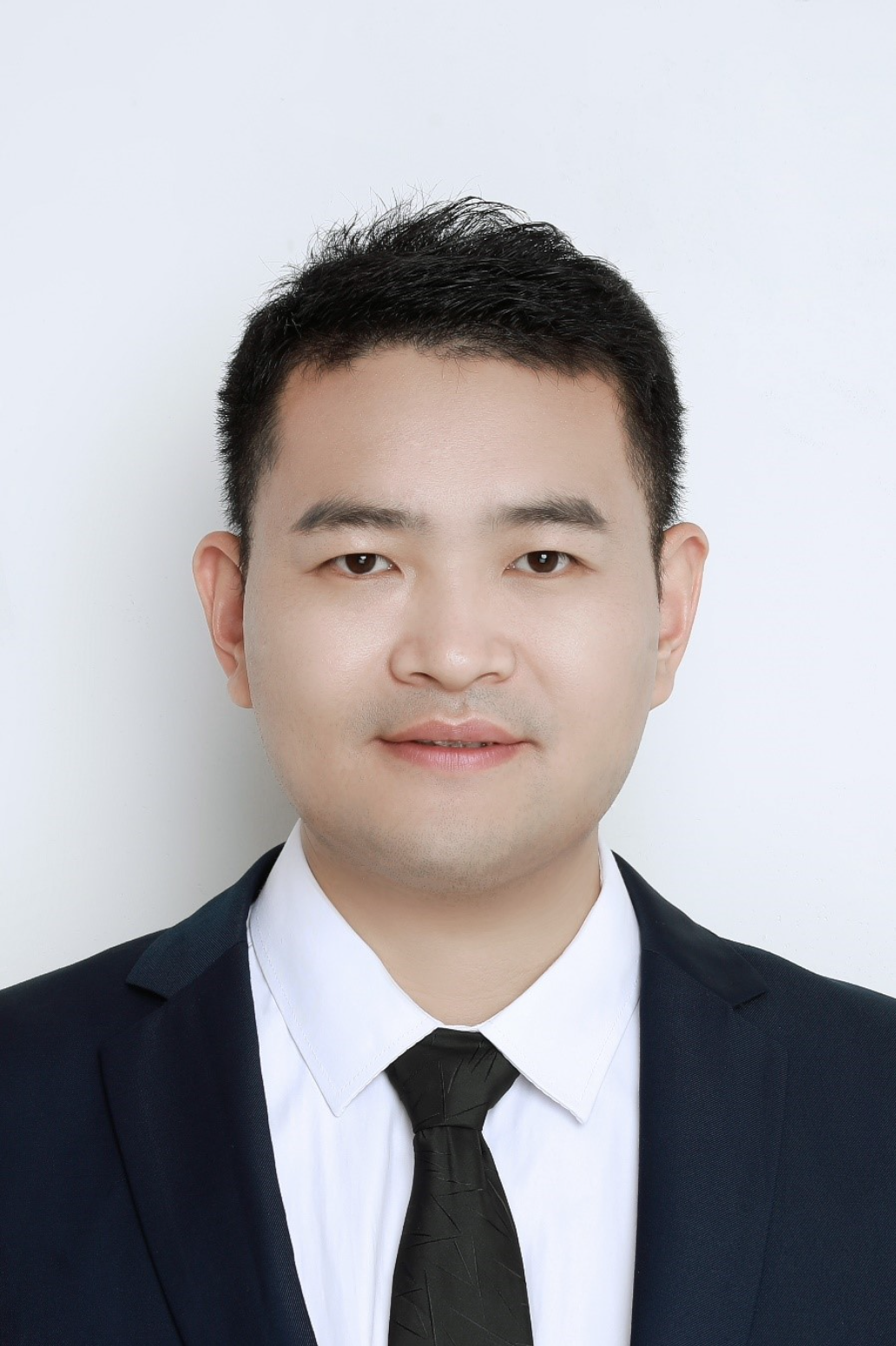}}]{Liqiang Nie}
Liqiang Nie, IAPR Fellow, is currently the dean with the School of Computer Science and Technology, Harbin Institute of Technology (Shenzhen). He received his B.Eng. and Ph.D. degree from Xi’an Jiaotong University and National University of Singapore (NUS), respectively. His research interests lie primarily in multimedia content analysis and information retrieval. Dr. Nie has co-/authored more than 100 CCF-A papers and 5 books, with 15k plus Google Scholar citations. He is an AE of IEEE TKDE, IEEE TMM, IEEE TCSVT, ACM ToMM, and Information Science. Meanwhile, he is the regular area chair or SPC of ACM MM, NeurIPS, IJCAI and AAAI. He is a member of ICME steering committee. He has received many awards over the past three years, like ACM MM and SIGIR best paper honorable mention in 2019, the AI 2000 most influential scholars 2020, SIGMM rising star in 2020, MIT TR35 China 2020, DAMO Academy Young Fellow in 2020, SIGIR best student paper in 2021, first price of the provincial science and technology progress award in 2021 (rank 1), and provincial youth science and technology award in 2022. Some of his research outputs have been integrated into the products of Alibaba, Kwai, and other listed companies.
\end{IEEEbiography}

\begin{IEEEbiography}[{\includegraphics[width=1in,height=1.25in,clip,keepaspectratio]{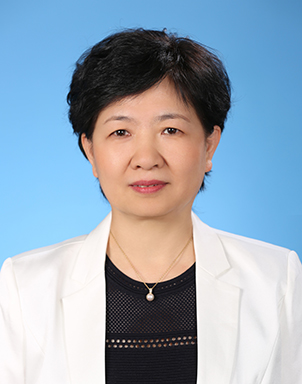}}]{Juanzi Li}
Juanzi Li is a Professor in Department of Computer Science and Technology at Tsinghua University. She received her Ph.D. degree from Tsinghua University in 2000. Her research interests include knowledge engineering and semantic web, text and social network mining. She has published over 100 research papers in top international journals and conferences such as TKDE, KDD, AAAI, ACL, EMNLP, etc.
\end{IEEEbiography}

% insert where needed to balance the two columns on the last page with
% biographies
%\newpage

% \begin{IEEEbiographynophoto}{Jane Doe}
% Biography text here.
% \end{IEEEbiographynophoto}

% You can push biographies down or up by placing
% a \vfill before or after them. The appropriate
% use of \vfill depends on what kind of text is
% on the last page and whether or not the columns
% are being equalized.

%\vfill

% Can be used to pull up biographies so that the bottom of the last one
% is flush with the other column.
%\enlargethispage{-5in}

% that's all folks
\end{document}